\documentclass[conference,a4paper]{IEEEtran}
\IEEEoverridecommandlockouts

\usepackage{cite}
\usepackage{amsmath,amssymb,amsfonts}
\usepackage{algorithmic}
\usepackage{graphicx}
\usepackage{textcomp}
\usepackage{xcolor}

\usepackage{placeins}
\usepackage{balance}

\usepackage{dblfloatfix}

\usepackage[hidelinks]{hyperref}

\def\BibTeX{{\rm B\kern-.05em{\sc i\kern-.025em b}\kern-.08em
    T\kern-.1667em\lower.7ex\hbox{E}\kern-.125emX}}

\newif\ifanonymous
\anonymousfalse 

\begin{document}

\title{A Comparative Study of Modern Object Detectors for Robust Apple Detection in Orchard Imagery}

\author{%
\ifanonymous
  \IEEEauthorblockN{Anonymous Authors}
  \IEEEauthorblockA{Affiliation omitted for double-blind review}
\else
  \begin{minipage}[t]{0.48\textwidth}
    \centering
    \textbf{Mohammed Asad}\\
    Dept. of Electronics and Communications Engineering\\
    Delhi Technological University\\
    New Delhi, India\\
    \texttt{mohammedasad\_ec22a17\_47@dtu.ac.in}
  \end{minipage}
  \hfill
  \begin{minipage}[t]{0.48\textwidth}
    \centering
    \textbf{Dr. Ajai Kumar Gautam}\\
    Dept. of Electronics and Communications Engineering\\
    Delhi Technological University\\
    New Delhi, India\\
    \texttt{ajai\_gautam@dce.ac.in}
  \end{minipage}\\[1.0ex]

  \begin{minipage}[t]{0.48\textwidth}
    \centering
    \textbf{Priyanshu Dhiman}\\
    Dept. of Electronics and Communications Engineering\\
    Delhi Technological University\\
    New Delhi, India\\
    \texttt{priyanshudhiman\_ec22a17\_71@dtu.ac.in}
  \end{minipage}
  \hfill
  \begin{minipage}[t]{0.48\textwidth}
    \centering
    \textbf{Rishi Raj Prajapati}\\
    Dept. of Electronics and Communications Engineering\\
    Delhi Technological University\\
    New Delhi, India\\
    \texttt{rishirajprajapati\_ee22b17\_60@dtu.ac.in}
  \end{minipage}
\fi
}

\maketitle

\begin{abstract}
For predicting yields, counting apples, robotic harvesting, and monitoring crops, it is very important to be able to accurately detect apples in orchard images. But changes in light, leaf clutter, thick fruit clusters, and partial occlusion often make performance worse. To provide a fair and reproducible comparison, this study establishes a controlled benchmark for single-class apple detection on the public AppleBBCH81 dataset using one deterministic train/validation/test split and a unified evaluation protocol across six representative detectors: YOLOv10n, YOLO11n, RT-DETR-L, Faster R-CNN (ResNet50-FPN), FCOS (ResNet50-FPN), and SSDLite320 (MobileNetV3-Large). Performance is evaluated primarily using COCO-style mAP@0.5 and mAP@0.5:0.95, and threshold-dependent behavior is further analyzed using precision-recall curves and fixed-threshold precision, recall, and F1-score at IoU = 0.5. On the validation split, YOLO11n achieves the best strict localization performance with mAP@0.5:0.95 = 0.6065 and mAP@0.5 = 0.9620, followed closely by RT-DETR-L (0.6012 and 0.9506) and YOLOv10n (0.5941 and 0.9501). At a fixed operating point with confidence $\geq 0.05$, YOLOv10n attains the highest F1-score (0.780870), whereas RT-DETR-L achieves very high recall (0.970218) but minimum precision (0.224605) because of many false positives at low confidence. These findings show that detector selection for orchard deployment should be guided not only by localization-aware accuracy but also by threshold robustness and the requirements of the downstream task.
\end{abstract}

\begin{IEEEkeywords}
Apple Detection, Precision Agriculture, Orchard Vision, AppleBBCH81 Dataset, Object Detection, YOLOv10, YOLO11, RT-DETR, Faster R-CNN, FCOS, SSDLite, COCO mAP, mAP@0.5:0.95, Precision-Recall Curve, Model Benchmarking, Real-Time Detection.
\end{IEEEkeywords}

\section{Purpose}
Apple detection is a basic task in precision agriculture. Its reliability impacts several applications, including yield estimation, fruit counting, robotic harvesting, thinning, and crop monitoring. However, in real orchard environments, detection is challenging due to significant variations in light, viewpoint, and fruit size. There are also a lot of leaves, branches, and shadows in the background. Apples often grow in thick clusters, which can make them hard to see. This makes it more likely that you will miss detections or get duplicates. These problems make it much harder to find orchards than to find controlled vision benchmarks. They also stress how important it is to test performance in real-world situations.

Numerous studies employing deep learning have demonstrated significant efficacy in fruit detection, particularly with apples. But it's hard to compare these studies because they often use different datasets, split methods, training resources, model types, and evaluation metrics. Specifically, studies that only report mAP@0.5 may conceal localization problems at stricter IoU thresholds and fail to elucidate how detector performance fluctuates with varying confidence levels. This limitation is important in orchards because systems that only count are prone to duplicate detections, and systems for harvesting need to be able to find things in messy places.

This paper introduces a controlled benchmark for single-class apple detection utilizing the AppleBBCH81 dataset, which includes bounding-box annotations in YOLO format. All tested models are trained and assessed utilizing a consistent training, validation, and testing division, accompanied by a standardized evaluation protocol. This approach ensures a fair comparison. The benchmark includes six representative detectors from various architectural families: YOLOv10n and YOLO11n as modern one-stage CNN baselines, RT-DETR-L as a real-time transformer detector, Faster R-CNN (ResNet50-FPN) as a two-stage baseline, FCOS (ResNet50-FPN) as an anchor-free one-stage baseline, and SSDLite320 (MobileNetV3-Large) as a lightweight deployment-oriented baseline.

This work has three main contributions. First, it sets up a reproducible and controlled benchmark for apple detection on AppleBBCH81, using one fixed split and one common evaluation pipeline for all models. Second, it compares six representative detector families under the same training and testing conditions, allowing for a fairer assessment between families than what has been available in previous orchard detection studies. Third, it demonstrates that detector ranking depends on the evaluation goal: YOLO11n scores highest under strict COCO-style localization metrics, while YOLOv10n achieves the best fixed-threshold F1-score, which is important for deployment-oriented decision-making.

The primary inquiry of this study is to determine which detector family offers the most dependable apple detection when employing stringent localization metrics in conjunction with threshold-dependent diagnostics. To solve this problem, the main metrics used are COCO-style mAP@0.5 and mAP@0.5:0.95, along with fixed-threshold precision, recall, F1-score, and precision-recall curves at IoU = 0.5. This evaluation design helps with both comparisons that take into account where the orchard is and practical analyses for putting the orchard in place.

The rest of this paper is set up like this. Section II talks about the workflow, how to prepare the dataset, how to preprocess it, how to train the model, and how to evaluate it. Section III presents the quantitative findings and comparative analysis. In Section IV, the main results are summarized, and ideas for future research are given.

\subsection{Models Used}
We evaluate six different types of object detectors for detecting apples as one class. We selected models from different architectural types in order to evaluate the localization performance, ability to reject poor detections at various threshold levels, and deployment capabilities in the same orchard environment.

\subsubsection{YOLOv10n}
YOLOv10n is a compact one-stage detection model designed for real-time object detection and is efficient in terms of computation by providing compact predictions that are well suited for deployment.\cite{wang2024yolov10}. In dense orchard scenes, precise confidence-threshold selection may be required to eliminate duplicate detections.

\subsubsection{YOLO11n}
YOLO11n is a lightweight Ultralytics one-stage detector to improve the accuracy-efficiency trade-off while maintaining real-time usability \cite{ultralytics2024yolo11docs}. Its multi-scale predictions are relevant for orchard visuals, where dimensions differ greatly based on perspective and distance.

\subsubsection{RT-DETR-L}
RT-DETR-L is a real-time transformer-based detector that uses a DETR-style established prediction framework instead of dense anchor creation \cite{zhao2024rtdetr}. By modeling global context, it may be good to handle cluttered scenes, as its practical operating-point behavior can differ with threshold selection.

\subsubsection{Faster R-CNN (ResNet50-FPN)}
Faster R-CNN is a two-step detector where a region proposal network creates potential regions that are subsequently classified and improved \cite{ren2015fasterrcnn,lin2017fpn}. It is considered a two-stage baseline because detection based on proposals frequently excels in situations with occlusion and dense clustering.

\subsubsection{FCOS (ResNet50-FPN)}
FCOS is an anchor-less one-stage detector that forecasts object positions directly from feature-map locations and employs centerness to minimize poor-quality predictions \cite{tian2019fcos}. It is included to investigate if anchor-free localization provides a benefit for small or partially hidden apples.

\subsubsection{SSDLite320 (MobileNetV3-Large)}
SSDLite320 is a lightweight SSD-based detector with a MobileNetV3 backbone that is developed for inference with fewer resources on mobile and edge devices\cite{liu2016ssd,howard2019mobilenetv3,torchvision_ssdlite320_mbv3_docs}. It is included as a deployment-oriented baseline to analyze the accuracy-efficiency trade-off in apple detection.

\subsection{Literature Review}
Fruit detection in orchards is a crucial element of precision farming as it aids in yield assessment, automated harvesting, thinning, and crop surveillance. Curation of visual benchmarks is a straightforward task. However, orchard photos provide unique challenges such as lighting variations as well as complicated backgrounds consisting of many leaves, branches, and skies surrounding and/or partially occluding clusters of fruit. These situations often result in overlooked detections of small or covered apples and repeated detections in densely populated areas.

\subsubsection{Fruit and Apple Detection in Orchard Environments}
Early orchard vision studies demonstrated that deep learning-based detectors can exceed the performance of handcrafted feature pipelines in field conditions. DeepFruits showed that deep neural networks can effectively detect fruits despite variations in the orchard \cite{sa2016deepfruits}. Bargoti and Underwood highlighted that clutter, occlusion, annotation quality, and diversity of datasets significantly impact performance in the field \cite{bargoti2017deep_fruit_detection}. These studies prompted the adoption of learning-based detectors in agricultural perception; however, they failed to establish a controlled benchmark among contemporary detector families.

\subsubsection{Datasets and Benchmarking}
The MinneApple project has created a reference dataset to allow sufficient comparison of apple counting and detection between different algorithms \cite{hani2020minneapple}. Some of the more recent research has evaluated UAV-based orchards for monitoring and yield estimation. \cite{kodors2024uav_ai_yield}. In this research, the apple at BBCH-stage 81 was selected as a point of comparison because it has a dataset publicly available in YOLO format, \cite{applebbch81_kaggle}, which means these studies can easily be reproduced in the future \cite{applebbch81_kaggle}.

\subsubsection{Apple-Specific Detection Improvements}
There have been multiple research projects focused on developing task-specific modifications for detecting apples in difficult outdoor environments, such as orchards. BFPNet is an example of a reported network that has been modified with a balanced pyramid network design (to improve the recall of small objects)\cite{sun2022bfpnet}. Several YOLO-based detection networks have also been modified to identify small apples and count them in difficult-to-detect conditions  \cite{ma2023yolov7tiny_apples}. Many multitask networks have extended the apple detection to include raw data estimation of other relevant orchard characteristics, such as estimating the size of apples \cite{ferrerferrer2023multitask_size}. Several augmentation methods have been examined; mosaic augmentation is considered effective in some orchard imaging \cite{kodors2024mosaic}.

\subsubsection{Modern Detection Architectures Relevant to This Study}
There are many different types of modern object detection, all of which have different tradeoffs in terms of speed, localization quality, and robustness. Two-stage detectors, such as Faster R-CNN, use a method of generating images that includes both proposals for items, as well as the ability to classify and refine those proposals once generated \cite{ren2015fasterrcnn}, while Feature Pyramid Network increases multi-scale representation and is used for small-object detection \cite{lin2017fpn}. ResNet backbones enhance feature extraction and stabilize training further \cite{he2016resnet}. One-stage detectors such as YOLO unify localization and classification for real-time inference \cite{redmon2016yolo}, and recent versions continue to enhance practical efficiency \cite{wang2024yolov10,ultralytics2024yolo11docs,ultralytics_repo}. FCOS provides an anchor-free alternative \cite{tian2019fcos}, while lightweight SSD-based detectors paired with MobileNet backbones target deployment on resource-constrained devices \cite{liu2016ssd,sandler2018mobilenetv2,howard2019mobilenetv3,torchvision_ssdlite320_mbv3_docs}. Transformer-based detection initiates in global attention-based context modeling \cite{vaswani2017attention,carion2020detr}, and RT-DETR adapts this family for real-time detection \cite{zhao2024rtdetr}.

\subsubsection{Comparative Summary of Prior Work}
Although prior studies have substantially advanced orchard fruit detection, many published results are difficult to compare directly because they use different datasets, task definitions, split strategies, detector families, and evaluation metrics. In particular, many studies focus on improving a specific architecture rather than comparing diverse detector families under a single protocol. Table~\ref{tab:lit_compare} summarizes representative prior work and highlights the gap addressed in this paper.

\begin{table*}[!t]
\centering
\caption{Summary of representative prior work on fruit and apple detection and their limitations relative to this study.}
\label{tab:lit_compare}
\setlength{\tabcolsep}{5pt}
\renewcommand{\arraystretch}{1.15}
\begin{tabular}{p{2.7cm} p{3.2cm} p{3.2cm} p{5.2cm}}
\hline
\textbf{Study} & \textbf{Main focus} & \textbf{Strength} & \textbf{Limitation relative to this study} \\
\hline
DeepFruits \cite{sa2016deepfruits} & Deep learning-based fruit detection in orchards & Demonstrated the feasibility of robust learning-based fruit detection in field imagery & Older detection setting; not a controlled benchmark across modern detector families \\
Bargoti and Underwood \cite{bargoti2017deep_fruit_detection} & Fruit detection under orchard clutter and occlusion & Highlighted the importance of real orchard conditions, dataset quality, and occlusion handling & Did not provide a unified cross-family benchmark under one evaluation protocol \\
MinneApple \cite{hani2020minneapple} & Apple detection and segmentation dataset & Enabled standardized evaluation on a public apple dataset & Focused on dataset creation; not a controlled comparison of multiple modern detector families on AppleBBCH81 \\
Hani et al. \cite{hani2020jfr_comparative} & Fruit detection versus counting performance & Showed that detection quality and counting quality are related but not identical & Focused on detection and counting objectives rather than detector-family benchmarking under one shared protocol \\
BFPNet \cite{sun2022bfpnet} & Small apple detection in complex orchard scenes & Improved small-object recall using a feature-pyramid design & Specialized architecture study rather than a broad detector-family comparison \\
Ma et al. \cite{ma2023yolov7tiny_apples} & Improved YOLO-based detection and counting of small apples & Strong task-specific optimization for challenging environments & Single-architecture focus; limited cross-family comparison \\
Kodors et al. \cite{kodors2024uav_ai_yield} & UAV-based orchard monitoring and yield estimation & Connected detection outputs to downstream agricultural applications & Different acquisition settings and objectives from controlled detector benchmarking \\
Kodors et al. \cite{kodors2024mosaic} & Data augmentation for agricultural imagery & Showed that augmentation choices influence agricultural detection performance & Focused on training strategy rather than standardized comparison of detector families \\
\hline
\end{tabular}
\end{table*}

\subsubsection{Positioning of the Present Study}
Motivated by these limitations, the present work benchmarks six representative detector families for single-class apple detection on AppleBBCH81 under one fixed deterministic split and one unified evaluation pipeline. We aim not to introduce a novel architecture but to provide a controlled and reproducible comparison to distinguish how contemporary, single-stage, dual-stage, anchor-less, transformer-based, and super-lightweight detectors produce results in orchard settings. This will differentiate this research from previous research, where only an individual architecture, a single change, a distinct dataset, or an alternative downstream effort was considered.
\section{Methodology}
\label{sec:method}

\subsection{Proposed Workflow}
\label{subsec:workflow}
\begin{figure}[t]
  \centering
  \includegraphics[width=\linewidth]{\detokenize{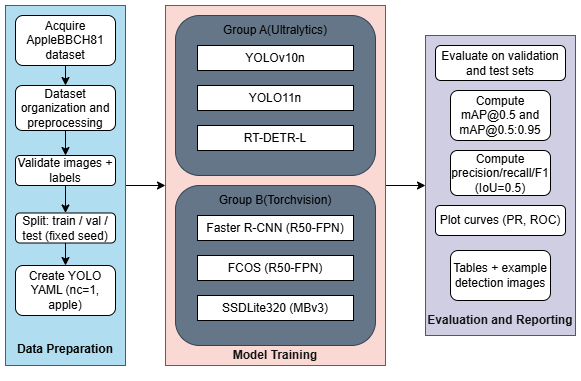}}
  \caption{Proposed workflow for apple detection on AppleBBCH81.}
  \label{fig:workflow}
\end{figure}

The proposed benchmark operates a controlled workflow to provide a fair comparison over detector families. First, the dataset is standardized, and at the image level, it is split into training, validation, and test subsets. Second, all models are instructed for the same single-class apple detection task by using clearly stated initialization and training settings. Third, predictions are estimated on the validation split using one common matching protocol and a shared set of localization-aware and threshold-dependent metrics. Finally, the trained detectors are measured using quantitative tables and diagnostic plots to examine both ranking quality and operating-point behavior. The complete benchmark pipeline is illustrated in Figure~\ref{fig:workflow}.

\subsection{Dataset and Split Protocol}
\label{subsec:dataset}
AppleBBCH81 is a public orchard dataset of RGB images annotated with apple bounding boxes in YOLO text format \cite{applebbch81_kaggle}. Each annotation line follows the format
\begin{center}
\texttt{<class\_id> <x\_center> <y\_center> <width> <height>}
\end{center}
where all coordinates are organized by image width and image height.

After preprocessing, the dataset contains 1,838 images and 15,309 apple bounding boxes. Among these images, 16 contain no apple annotations and are retained as valid negative samples. A fixed image-level split is used for all experiments to ensure direct comparability across models: 1,470 training images with 12,256 boxes, 183 validation images with 1,511 boxes, and 185 test images with 1,542 boxes. This corresponds approximately to an 80/10/10 split. All quantitative results reported in this paper are computed on the validation split.

\subsection{Experimental Setup}
\label{subsec:setup}
All experiments were executed in Google Colab using an NVIDIA A100-SXM4-40GB GPU, Python 3.12.12, PyTorch 2.10.0+cu128, Torchvision 0.25.0+cu128, and CUDA 12.8. YOLOv10n, YOLO11n, and RT-DETR-L were trained and evaluated using Ultralytics, whereas Faster R-CNN, FCOS, and SSDLite320 were trained and evaluated using PyTorch and Torchvision. Randomness was controlled through deterministic seeding, and the Ultralytics run configuration files record \texttt{seed = 0}. Dataset integrity checks confirmed that all images were readable, each image had a label file, and no samples required exclusion or correction. Empty-label images were retained throughout training and evaluation to reduce false-positive bias. Table~\ref{tab:exp_setup} summarizes the main training configurations used for all six detectors.

\begin{table*}[!t]
\centering
\caption{Summary of model training configurations used in the benchmark.}
\label{tab:exp_setup}
\setlength{\tabcolsep}{5pt}
\renewcommand{\arraystretch}{1.15}
\begin{tabular}{l l l c c c l c}
\hline
\textbf{Model} & \textbf{Framework} & \textbf{Initialization} & \textbf{Input size} & \textbf{Batch} & \textbf{Epochs} & \textbf{Optimizer} & \textbf{Learning rate} \\
\hline
YOLOv10n & Ultralytics & COCO pretrained & 640 & 16 & 50 & auto & 0.01 \\
YOLO11n & Ultralytics & COCO pretrained & 640 & 16 & 50 & auto & 0.01 \\
RT-DETR-L & Ultralytics & COCO pretrained & 640 & 16 & 50 & auto & 0.01 \\
Faster R-CNN (R50-FPN) & Torchvision & COCO pretrained & 800 to 1333 & 4 & 10 & AdamW & $1\times10^{-4}$ \\
FCOS (R50-FPN) & Torchvision & COCO pretrained & 800 to 1333 & 4 & 10 & AdamW & $1\times10^{-4}$ \\
SSDLite320 (MobileNetV3-Large) & Torchvision & COCO pretrained & 320 & 4 & 10 & AdamW & $1\times10^{-4}$ \\
\hline
\end{tabular}
\end{table*}

\subsection{Data Standardization and Preprocessing}
\label{subsec:preprocess}
To maintain a fair comparison, all detectors use the same fixed split described in Section~\ref{subsec:dataset}. Preprocessing is restricted to label conversion, consistent handling of negative images, and framework-specific input preparation.

First, the task is formulated as single-class apple detection. For Ultralytics, Apple is assigned class ID 0. For Torchvision models, Apple is assigned the class label 1 because the label 0 is reserved for the background. Second, YOLO labels of the form $(x_c, y_c, w, h)$ are converted to absolute pixel coordinates $(x_1, y_1, x_2, y_2)$ for Torchvision targets, using the corresponding image width and height. Third, negative images are preserved by keeping empty label files and passing empty target tensors to Torchvision models.

Input resizing and normalization follow the conventions of each framework. YOLOv10n, YOLO11n, and RT-DETR-L use a fixed input resolution of $640\times640$ with Ultralytics letterboxing. Faster R-CNN and FCOS use the Torchvision resize policy with \texttt{min\_size} set to 800 and \texttt{max\_size} set to 1333, together with ImageNet normalization mean $[0.485, 0.456, 0.406]$ and standard deviation $[0.229, 0.224, 0.225]$. SSDLite320 uses a fixed input size of 320 with normalization mean $[0.5, 0.5, 0.5]$ and standard deviation $[0.5, 0.5, 0.5]$.

For data augmentation, Ultralytics models use the training settings recorded in the run configuration files: \texttt{hsv\_h}=0.015, \texttt{hsv\_s}=0.7, \texttt{hsv\_v}=0.4, \texttt{degrees}=0.0, \texttt{translate}=0.1, \texttt{scale}=0.5, \texttt{shear}=0.0, \texttt{perspective}=0.0, \texttt{flipud}=0.0, \texttt{fliplr}=0.5, \texttt{mosaic}=1.0, \texttt{mixup}=0.0, \texttt{copy\_paste}=0.0, and \texttt{erasing}=0.4. Torchvision models do not use mosaic or mixup and are trained with tensor conversion and the model-specific normalization described above.

\subsection{Model Training}
\label{subsec:training}
All six detectors are initialized from COCO-pretrained weights and fine-tuned for single-class apple detection using the same training split and the same validation split. For the Ultralytics models, YOLOv10n, YOLO11n, and RT-DETR-L are trained with \texttt{imgsz}=640, \texttt{epochs}=50, \texttt{batch}=16, \texttt{optimizer=auto}, base learning rate \texttt{lr0}=0.01, \texttt{weight\_decay}=0.0005, and \texttt{seed}=0. Initial checkpoints are \texttt{yolov10n.pt}, \texttt{yolo11n.pt}, and \texttt{rtdetr-l.pt}. Reported Ultralytics results correspond to the best checkpoint picked by validation mAP@0.5:0.95 within each run.

For the Torchvision models, Faster R-CNN (ResNet50-FPN), FCOS (ResNet50-FPN), and SSDLite320 (MobileNetV3-Large) are fine-tuned using a batch size of 4 for 10 epochs with the AdamW optimizer and learning rate $1\times10^{-4}$. Faster R-CNN and FCOS use the Torchvision resize policy of 800 to 1333 pixels with ImageNet normalization, whereas SSDLite320 uses a fixed input size of 320 with normalization mean and standard deviation equal to 0.5 in each channel.

\subsection{Evaluation Protocol}
\label{subsec:metrics}
All reported metrics are computed on the validation split. The evaluation protocol is designed to compare detectors using both strict localization quality and threshold-dependent deployment behavior.

\subsubsection{Intersection over Union}
For a prediction box $B_p$, a ground-truth box $B_{gt}$, the intersection over union is defined as
\begin{equation}
\mathrm{IoU} = \frac{|B_p \cap B_{gt}|}{|B_p \cup B_{gt}|}.
\label{eq:iou}
\end{equation}

\subsubsection{One-to-One Matching Rule}
To compute object-level true positives, false positives, and false negatives at a given confidence threshold, predictions are first sorted by confidence in descending order. Each prediction goes with the best unmatched ground-truth box. If the best IoU is at least 0.5 and the corresponding ground-truth box has not been matched previously, the prediction is calculated as a true positive. On the other hand, it is counted as a false positive. Any unmatched ground-truth boxes are calculated as false negatives. This one-to-one matching rule prevents multiple detections from being counted as correct for the same apple.

\subsubsection{COCO-style mAP Metrics}
The primary localization-aware metrics are mAP@0.5 and mAP@0.5:0.95. The former reports average precision at IoU = 0.5, whereas the latter averages average precision over IoU thresholds from 0.50 to 0.95 in increments of 0.05. These metrics are used as the primary basis for comparing detector quality because they evaluate both detection ranking and localization precision over a confidence sweep.

\subsubsection{Fixed-Threshold Precision, Recall, and F1-score}
To examine deployment-oriented operating-point behavior, precision, recall, and F1-score are computed at IoU = 0.5 and confidence threshold $\ge 0.05$:
\begin{equation}
\mathrm{Precision} = \frac{TP}{TP + FP}, \qquad
\mathrm{Recall} = \frac{TP}{TP + FN},
\label{eq:pr}
\end{equation}
\begin{equation}
\mathrm{F1} = 2 \cdot \frac{\mathrm{Precision}\cdot\mathrm{Recall}}{\mathrm{Precision} + \mathrm{Recall}}.
\label{eq:f1}
\end{equation}
These metrics are particularly relevant for orchard applications such as counting and robotic harvesting, where the practical operating threshold directly affects duplicate detections and missed fruit.

\subsubsection{Precision-Recall Curves and PR AUC}
Precision-recall curves are generated by sweeping the confidence threshold and plotting precision against recall using the one-to-one matching rule at IoU $\ge 0.5$. PR AUC is computed by trapezoidal integration over recall. In this study, precision-recall analysis is treated as the main threshold-dependent diagnostic because it is more appropriate for object detection than ROC-based analysis.

\subsection{Statistical Uncertainty and Reporting}
\label{subsec:stats}
To strengthen the interpretation of fixed-threshold results, 95\% Wilson confidence intervals are used for precision and recall. For precision, the number of trials is taken as $TP + FP$, and for recall, the number of trials is taken as $TP + FN$. These intervals are reported only for the fixed-threshold operating-point metrics and are intended to provide uncertainty estimates for the observed precision and recall values on the validation split.

The final comparison, therefore, includes three components: COCO-style mAP@0.5 and mAP@0.5:0.95 as the primary localization-aware metrics, precision-recall curves and PR AUC as threshold-dependent diagnostics, and object-level fixed-threshold precision, recall, and F1-score with TP, FP, and FN counts. Because true negatives are not well-defined at the object level in detection tasks, confusion-style reporting is restricted to TP, FP, and FN.

\begin{figure}[t]
  \centering
  \includegraphics[width=\linewidth]{\detokenize{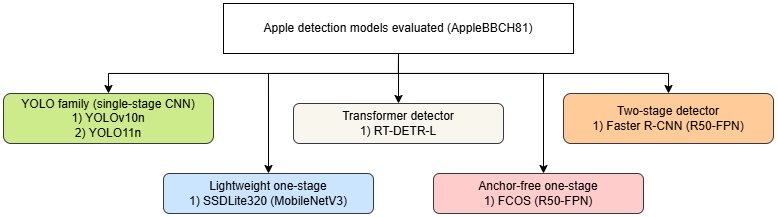}}
  \caption{Model taxonomy of the evaluated detectors.}
  \label{fig:taxonomy}
\end{figure}

Figure~\ref{fig:taxonomy} groups the evaluated detectors by architectural family and highlights the diversity of one-stage, two-stage, anchor-free, transformer-based, and lightweight models included in the benchmark.

\FloatBarrier
\section{Results \& Findings}
\label{sec:results}

\subsection{Quantitative Results (Validation Set)}
All results analyzed in this section are computed on the validation split, which contains 183 images and 1,511 ground-truth apple instances. Table~\ref{tab:val_map} presents the primary COCO-style localization-aware metrics, namely mAP@0.5 and mAP@0.5:0.95. Table~\ref{tab:val_diag} complements these outcomes with fixed-threshold operating-point diagnostics at IoU $=0.5$ and confidence $\ge 0.05$, including TP, FP, FN, precision, recall, F1-score, PR AUC, and 95\% Wilson confidence intervals for precision and recall. Figures~\ref{fig:pr_curve} and \ref{fig:conf_summary} further illustrate threshold behavior and the low-threshold error profile of the best-mAP model.

\begin{table}[!htbp]
\centering
\caption{COCO-style validation results for single-class apple detection.}
\label{tab:val_map}
\setlength{\tabcolsep}{4pt}
\renewcommand{\arraystretch}{1.15}
\resizebox{\linewidth}{!}{%
\begin{tabular}{l l c c}
\hline
\textbf{Family} & \textbf{Model} & \textbf{mAP@0.5:0.95} & \textbf{mAP@0.5} \\
\hline
Ultralytics & YOLO11n & 0.6065 & 0.9620 \\
Ultralytics & RT-DETR-L & 0.6012 & 0.9506 \\
Ultralytics & YOLOv10n & 0.5941 & 0.9501 \\
Torchvision & Faster R-CNN (R50-FPN) & 0.5278 & 0.9091 \\
Torchvision & FCOS (R50-FPN) & 0.4926 & 0.8993 \\
Torchvision & SSDLite320 (MobileNetV3-Large) & 0.2125 & 0.5400 \\
\hline
\end{tabular}%
}
\end{table}

Table~\ref{tab:val_map} shows that YOLO11n achieves the best strict localization-aware performance on the validation split, with mAP@0.5:0.95 of 0.6065 and mAP@0.5 of 0.9620. RT-DETR-L and YOLOv10n follow closely, with only a narrow gap separating the top three models under COCO-style evaluation. Among the Torchvision baselines, Faster R-CNN outperforms FCOS, while SSDLite320 is substantially lower, which is consistent with its lightweight design and reduced input resolution. The findings suggest that contemporary real-time detectors deliver the highest level of localization accuracy within the controlled benchmark framework applied in this research.

\subsection{Operating-Point Analysis and Threshold Behavior}
Table~\ref{tab:val_diag} reports present fixed-threshold operating-point statistics calculated through one-to-one greedy matching at an Intersection over Union (IoU) of 0.5 and a confidence level of at least 0.05. This table is designed to enhance mAP-based ranking by illustrating the performance of the detectors at a particular deployment-focused threshold instead of across a comprehensive range of confidence levels.

\begin{table*}[!t]
\centering
\caption{Validation diagnostics at IoU $=0.5$ and confidence $\ge 0.05$ using one-to-one greedy matching.}
\label{tab:val_diag}
\setlength{\tabcolsep}{4pt}
\renewcommand{\arraystretch}{1.15}
\begin{tabular}{l c c c c c c c c c}
\hline
\textbf{Model} & \textbf{TP} & \textbf{FP} & \textbf{FN} & \textbf{PR AUC} & \textbf{Precision} & \textbf{95\% CI} & \textbf{Recall} & \textbf{95\% CI} & \textbf{F1} \\
\hline
YOLO11n & 1465 & 920 & 46 & 0.958 & 0.614256 & [0.594551, 0.633593] & 0.969557 & [0.959632, 0.977099] & 0.752053 \\
RT-DETR-L & 1466 & 5061 & 45 & 0.950 & 0.224605 & [0.214645, 0.234890] & 0.970218 & [0.960383, 0.977669] & 0.364767 \\
YOLOv10n & 1445 & 745 & 66 & 0.947 & 0.659817 & [0.639710, 0.679365] & 0.956320 & [0.944806, 0.965520] & 0.780870 \\
Faster R-CNN (R50-FPN) & 1391 & 717 & 120 & 0.900 & 0.659867 & [0.639368, 0.679784] & 0.920582 & [0.905858, 0.933174] & 0.768721 \\
FCOS (R50-FPN) & 1417 & 1289 & 94 & 0.904 & 0.523651 & [0.504813, 0.542422] & 0.937790 & [0.924465, 0.948893] & 0.672042 \\
SSDLite320 (MobileNetV3-Large) & 1323 & 30213 & 188 & 0.542 & 0.041952 & [0.039795, 0.044221] & 0.875579 & [0.857978, 0.891275] & 0.080068 \\
\hline
\end{tabular}
\end{table*}

At this specific operating point, YOLOv10n shows the best balance between precision and recall, achieving the highest F1-score of 0.780870 with TP = 1445, FP = 745, and FN = 66. Faster R-CNN also performs competitively at this threshold, while YOLO11n, despite having the highest mAP, yields more false positives than YOLOv10n and, as a result, has a lower fixed-threshold F1-score of 0.752053. This distinction indicates that a detector with superior ranking quality across a confidence sweep is not necessarily the one that shows the best performance at a low threshold.

RT-DETR-L clearly explains this difference. While it gains impressive COCO-style mAP and a high PR AUC of 0.950, it produces 5,061 false positives at a confidence level of $\ge 0.05$, leading to a precision of 0.224605 despite having a very high recall of 0.970218. These key aspects indicate that RT-DETR-L is capable of effectively ranking true detections across various thresholds, but it still demands careful threshold adjustments to minimize the occurrence of excessive duplicates and false positives in practical applications like fruit counting.

The 95\% Wilson confidence intervals in Table~\ref{tab:val_diag} improve the fixed-threshold interpretation by refining the uncertainty surrounding the measured precision and recall metrics in the validation set. For example, YOLOv10n and Faster R-CNN have similar precision ranges, while YOLO11n and RT-DETR-L illustrate higher recall ranges but remain lower precision compared to YOLOv10n. These ranges do not substitute for all significance testing of mAP, yet they offer a statistically sound estimate of uncertainty for the recorded operating-point outcomes.

\begin{figure}[!htbp]
  \centering
  \includegraphics[width=\linewidth]{\detokenize{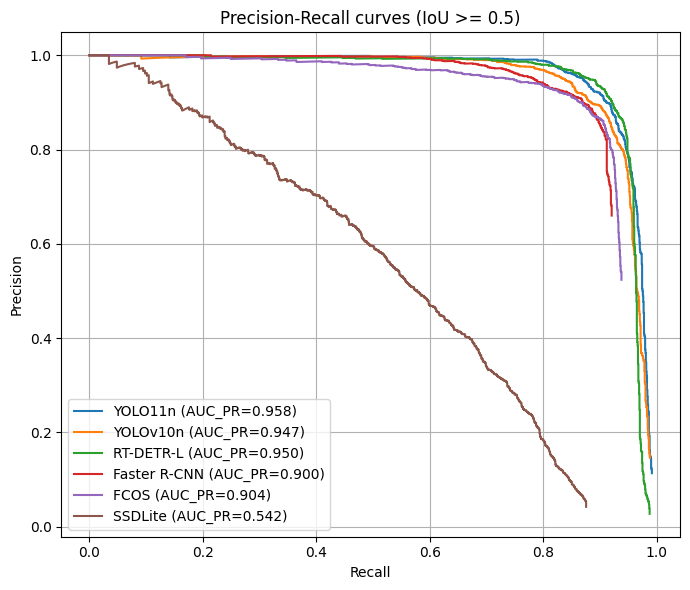}}
  \caption{Precision-recall curves for all models on the validation split at IoU $=0.5$.}
  \label{fig:pr_curve}
\end{figure}

Figure~\ref{fig:pr_curve} gives the main threshold-dependent comparison across models. YOLO11n, RT-DETR-L, and YOLOv10n maintain high precision over most of the recall range, which is consistent with their strong mAP and PR AUC values. In contrast, SSDLite320 deteriorates much earlier as recall increases, indicating that its detections become substantially less reliable when the model is pushed toward high recall. This figure, therefore, explains why PR AUC is a useful complement to mAP when comparing the threshold robustness.

\begin{figure}[!htbp]
  \centering
  \includegraphics[width=0.92\linewidth,height=0.30\textheight,keepaspectratio]{\detokenize{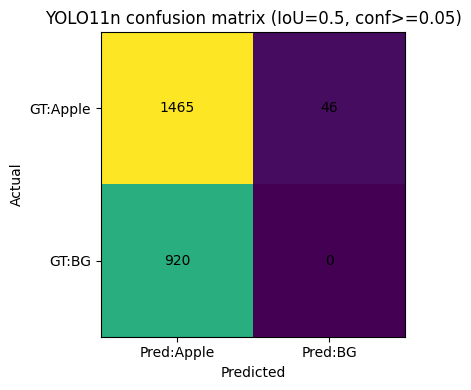}}
  \caption{Object-level confusion summary for YOLO11n on the validation split at IoU $=0.5$ and confidence $\ge 0.05$ (TP = 1465, FP = 920, FN = 46; TN is not defined for object detection).}
  \label{fig:conf_summary}
\end{figure}

Figure~\ref{fig:conf_summary} clarifies the low-threshold error profile of YOLO11n, which is the best model by mAP@0.5:0.95. Although the model achieves the highest strict localization performance, it still produces 920 false positives at the selected operating point. This explains why YOLO11n does not achieve the highest fixed-threshold F1-score despite leading the benchmark in COCO-style evaluation.

\subsection{Implications for Real-World Orchard Deployment}
On the other hand, aggregate benchmark scores and the practical orchard deployment, introduces some additional constraints. First, fruit counting and yield estimation are sensitive to duplicate detections, so a detector with high recall but many false positives is not the best choice in practice. Secondly, robotic harvesting involves reliable localization regardless of occlusions, dense clustering, and varying lighting conditions, highlighting the importance of both localization precision and threshold resilience. Third, real-world deployment involves embedded or edge hardware, where model size, memory use, and latency may favor lightweight architectures even when their strict mAP is lower. Finally, generalization across orchards, cultivars, seasons, and camera viewpoints remains an important challenge that is not fully captured by a single-dataset benchmark. These considerations indicate that detector selection should be matched to the target agricultural application rather than based on a single headline metric alone.

\FloatBarrier

\section{Conclusion \& Future Scope}

This paper presented a controlled benchmark of six modern object detectors for single-class apple detection on AppleBBCH81 using one fixed train/validation/test split and a unified training and evaluation protocol. Under strict COCO-style localization-aware evaluation on the validation split, YOLO11n achieved the best overall performance with mAP@0.5:0.95 of 0.6065 and mAP@0.5 of 0.9620, where RT-DETR-L and YOLOv10n followed closely. However, detector ranking changed at a deployment-oriented operating point of IoU $=0.5$ and confidence $\ge 0.05$, where YOLOv10n achieved the highest F1-score of 0.780870, showing the best balance between precision and recall at a low threshold. RT-DETR-L also showed strong ranking quality over threshold sweeps but produced many false positives at the selected operating point, demonstrating that high mAP alone does not guarantee the most stable practical behavior for orchard deployment. Overall, the results show that detector selection for orchard applications should be analyzed with strict localization quality, threshold robustness, and the needs of the downstream task, mainly in settings such as fruit counting and robotic harvesting that are sensitive to duplicate detections and missed fruit. Future research should scale this benchmark to include cross-orchard and cross-season assessments, examine confidence calibration and the selection of uncertainty-aware thresholds, study counting-aware and occlusion-aware training methods, and analyze deployment trade-offs, such as inference speed, model size, and efficiency on edge devices.


\section*{Acknowledgements}
Experiments were conducted using Google Colab for training and evaluation. The AppleBBCH81 dataset was obtained from Kaggle and is gratefully acknowledged.

\FloatBarrier
\balance

\bibliographystyle{IEEEtran}
\bibliography{references}

\end{document}